\documentclass{article} 
\pdfoutput=1
\usepackage{nips13submit_e,times}

\usepackage{url}
\usepackage{bbm}
\usepackage{amsmath,amsfonts,amssymb}
\usepackage{amsthm}
\usepackage{graphicx} 
\usepackage{subfigure}

\usepackage{algorithm}
\usepackage{algorithmic}

\usepackage{authblk}

\newcommand\blfootnote[1]{%
  \begingroup
  \renewcommand\thefootnote{}\footnote{#1}%
  \addtocounter{footnote}{-1}%
  \endgroup
}

\usepackage{hyperref}
\usepackage[square,sort,comma,numbers]{natbib}

\newtheorem{theorem}{Theorem}

\title{Stochastic Variational Inference for Hidden Markov Models}

\author[]{\textbf{Nicholas J. Foti$^\dagger$}}
\author[]{\textbf{Jason Xu$^\dagger$}}
\author[]{\textbf{Dillon Laird}}
\author[]{\textbf{Emily B. Fox}}
\affil[]{\vspace{-0.1in}University of Washington \texttt{\small \{nfoti@stat,jasonxu@stat,dillonl2@cs,ebfox@stat\}.washington.edu}}

\newcommand{\Ell}{\mathcal{L}}
\newcommand{\mb}{\mathbf}
\newcommand{\ind}{\mathbbm{1}}
\newcommand{\E}{\mathbbm{E}}
\newcommand{\norm}[1]{\left | \left | #1 \right | \right |}
\newcommand{\reals}{\mathbb{R}}
\newcommand{\N}{\mathrm{N}}
\newcommand{\Dir}{\mathrm{Dir}}
\newcommand{\NIW}{\mathrm{NIW}}

\newcommand{\Mult}{\mathrm{Mult}}

\DeclareMathOperator*{\argmax}{arg\,max}

\nipsfinalcopy 

\begin{document}

\maketitle

\begin{abstract}
Variational inference algorithms have proven successful for Bayesian analysis in large data settings, with recent advances using stochastic variational inference~(SVI).
However, such methods have largely been studied in independent or exchangeable data settings.  We develop an SVI algorithm to learn the parameters of hidden Markov models (HMMs) in a time-dependent data setting. The challenge in applying stochastic optimization in this setting arises from dependencies in the chain, which must be broken to consider minibatches of observations.
We propose an algorithm that harnesses the memory decay of the chain to adaptively bound errors arising from edge effects.
 We demonstrate the effectiveness of our algorithm on synthetic experiments and a large genomics dataset 
 where a batch algorithm is computationally infeasible. \blfootnote{$^\dagger$ Co-first authors contributed equally to this work.}

\end{abstract} 

\section{Introduction}
\label{Intro}

Modern data analysis has seen an explosion in the size of the datasets
available to analyze.  Significant progress has been made scaling
machine learning algorithms  to these massive
datasets based on optimization procedures \citep{Robbins:Monro:1951,Bottou:1998,Bottou:2010}.
For example, stochastic gradient descent employs noisy estimates
of the gradient based on \textit{minibatches} of data, avoiding
a costly gradient computation using the full dataset \cite{Nemirovski:2009}.  There is
considerable interest in leveraging these methods for Bayesian
inference since traditional algorithms such as Markov chain Monte
Carlo (MCMC) scale poorly to large datasets, though subset-based MCMC methods have been recently proposed
as well \citep{Welling:Teh:2011,Maclaurin:2014,Wang:Dunson:2014,Neiswanger:Xing:2014}.

\emph{Variational Bayes} (VB) casts posterior inference as a tractable optimization
problem by minimizing the Kullback-Leibler divergence between
the target posterior and a family of simpler \emph{variational
distributions}.  Thus, VB provides a natural framework to incorporate
ideas from stochastic optimization to perform scalable Bayesian inference.
Indeed, a scalable modification to VB harnessing stochastic
gradients---\emph{stochastic variational inference} (SVI)---has recently been applied to a variety of Bayesian latent variable models
\citep{Hoffman:2013,Bryant:2012}.  Minibatch-based
VB methods have also proven effective in a streaming setting
where data arrives sequentially
\cite{Broderick:Boyd:Wibisono:Wilson:Jordan:2013}.  

However, these algorithms have been developed assuming independent or exchangeable data. 
One exception is the SVI algorithm for the mixed-membership stochastic
block model~\cite{Gopalan:Mimno:Gerrish:Freedman:Blei:2012}, but independence at the level of the generative model must be exploited.  
SVI for Bayesian time series including HMMs was recently considered in settings where each minibatch is a set of \emph{independent} series   \cite{Johnson:Willsky:2014}, 
though in this setting again dependencies do not need to be broken.



In contrast, we are interested in applying SVI to very long time series.
As a motivating example, consider the application in Sec.~\ref{sec:exper} of a genomics dataset consisting of $T=250$ million observations in $12$ dimensions modeled via an HMM to learn human chromatin structure.  An analysis of the entire sequence is computationally prohibitive using standard Bayesian inference techniques for HMMs due to a per-iteration complexity linear in $T$. Unfortunately,
despite the simple chain-based dependence structure, applying a minibatch-based method is not obvious. In particular, there are two potential issues immediately arising in sampling subchains as minibatches:  (1) the subsequences are not mutually independent, and (2) updating the latent variables in the subchain ignores the data outside of the subchain introducing error.
We show that for (1), appropriately scaling the noisy subchain gradients preserves unbiased gradient estimates. To address (2), we propose an approximate message-passing scheme that adaptively bounds error by accounting for memory decay of the chain.  


%


We prove that our proposed \emph{SVIHMM} algorithm converges to a local mode of the batch objective, and empirically demonstrate similar performance to batch VB in significantly less time on synthetic datasets.
We then consider our genomics application 
and show that SVIHMM allows efficient Bayesian inference on this massive dataset where batch inference is computationally infeasible.

\section{Background} 
\subsection{Hidden Markov models}

Hidden Markov models (HMMs) \cite{Rabiner:1989} are a class of discrete-time doubly
stochastic processes consisting of observations $y_t$ and latent states $x_t \in
\{1,\dots,K\}$ generated by a discrete-valued Markov chain. Specifically, for $\mb{y} = (y_1, \ldots,
y_T)$ and $\mb{x} = (x_1, \ldots, x_T)$, the joint distribution factorizes as
\begin{equation} \label{eq:HMMprob}
p( \mb{x, y}) =  \mb{\pi}_0(x_1) p(y_1|x_1) \prod_{t=2}^T p(x_t| x_{t-1}, A) p(y_t|x_t, \phi) 
\end{equation}
where $A = \left [ A_{ij} \right ]_{i,j=1}^K$ is the \textit{transition matrix} with
$A_{ij} = \text{Pr}(x_t = j| x_{t-1} = i)$, $\phi = \{\phi_k\}_{k=1}^K$ the
\textit{emission parameters}, and $ \pi_0$ the \textit{initial
distribution}.  We denote the set of HMM
parameters as $\mb{\theta} = (\pi_0, A, \phi)$.  
We assume that the underlying chain is irreducible and aperiodic so that a
\textit{stationary distribution} $\pi$ exists and is unique.  Furthermore, we assume
that we observe the sequence at stationarity so that $\pi_0=\pi$, where $\pi$
is given by the leading left-eigenvector of $A$.  As such, we do not seek to
learn $\pi_0$ in the setting of observing a single realization of a long chain.

We specify conjugate Dirichlet priors on the rows of the transition matrix as
\begin{align}
    p(A) = \prod_{j=1}^K \text{Dir} (A_{i:} \mid \alpha^A_{j} ).
\end{align}
Here, $\mbox{Dir}(\pi\mid \alpha)$ denotes a $K$-dimensional Dirichlet
distribution with concentration parameters $\alpha$.  Although our methods are more
broadly applicable, we focus on HMMs with
multivariate Gaussian emissions where $\phi_k = \{\mu_k,\Sigma_k\}$, with conjugate normal-inverse-Wishart (NIW) prior
\begin{align}
y_t \mid x_t \sim N( y_t \mid \mu_{x_t}, \Sigma_{x_t}), \hspace{16pt}
\phi_k = (\mu_k, \Sigma_k)  \sim \text{NIW}(\mu_0, \kappa_0, \Sigma_0, \nu_0).
\end{align}
For simplicity, we suppress dependence on $\theta$ and write
$\pi(x_0)$, $p(x_t|x_{t-1})$, and $p(y_t|x_t)$ throughout.


\subsection{Structured mean-field VB for HMMs}
\label{sec:structuredMF}

We are
interested in the \textit{posterior distribution} of the state sequence and
parameters given an observation sequence, denoted $p(\mb{x}, \theta |
\mb{y})$.  While evaluating marginal likelihoods, $p(\mb{y}|\theta)$, and
most probable state sequences, $\argmax_{\mb{x}}
p(\mb{x} | \mb{y}, \theta)$, are tractable via 
the \text{forward-backward} (FB) algorithm when parameter values $\theta$ are \emph{fixed}~\cite{Rabiner:1989},
exact computation of the posterior is intractable for HMMs.
Markov chain Monte Carlo (MCMC) provides a widely used sampling-based approach to posterior inference in HMMs~\cite{FruhwirthSchnatter:2006,Scott:2002}.
We instead focus on variational Bayes (VB), an optimization-based approach that
approximates $p(\mb{x}, \theta | \mb{y})$ by a variational distribution $q(\theta, \mb{x})$ within a simpler family.
Typically, for HMMs a \emph{structured mean field} approximation is considered: 
\begin{align}
q(\theta, \mb{x}) = q(A) q(\phi) q(\mb{x}),
\label{eq:structuredMF}
\end{align}
breaking dependencies only between the parameters $\theta = \left\{A, \phi \right\}$ and latent state sequence $\mb{x}$ \cite{Beale:2003}. 
Note that making a full mean field assumption in which $q (\mb{x}) = \prod_{i=1}^T q(x_i)$
loses crucial information about the latent chain needed for accurate inference. 

Each factor in Eq.~\eqref{eq:structuredMF} is endowed with its own
variational parameter and is set to be in the same exponential family
distribution as its respective complete conditional. 
The variational parameters are optimized to maximize the \textit{evidence
lower bound} (ELBO) $\Ell$: 
\begin{equation}\label{eq:elbo}
\ln p( \mb{y})  \geq E_{q} \left[ \ln p(\theta) \right] - E_{q} \left[ \ln q(\theta) \right] +  E_{q} \left[ \ln p( \mb{y}, \mb{x} | \theta) \right] - E_{q} \left[ \ln q( \mb{x} ) \right] := \Ell(q(\theta), q(\mb{x})) .
%
\end{equation}

Maximizing $\Ell$ is equivalent to minimizing the KL divergence
$\text{KL}(q(\mb{x},\theta)||p(\mb{x},\theta | \mb{y}))$ \cite{Jordan:1999}.  In practice, we
alternate updating the \emph{global parameters} $\theta$---those coupled to
the entire set of observations---and the \emph{local variables}
$\{x_t\}$---a variable corresponding to each observation, $y_t$. Details on computing the terms in the equations and algorithms that follow are in the Supplement.

The \textit{global update} is derived by differentiating $\Ell$ with respect
to the global variational parameters \cite{Beale:2003}.
Assuming a conjugate exponential family leads to a simple coordinate ascent update~\cite{Hoffman:2013}: 
\begin{align}
    \label{eq:globalstep}
\mb{w} = \mb{u} + E_{q(\mb{x})} \left[ t(\mb{x,y}) \right].
\end{align}
Here, $t(\mb{x,y})$ denotes the vector of sufficient statistics, and $\mb{w} =
(\mb{w}^A, \mb{w}^\phi)$  and $\mb{u}= (\mb{u}^A, \mb{u}^\phi)$ the variational parameters and model hyperparameters, respectively, in natural parameter form. 

The \textit{local update} is derived analogously, yielding the optimal
variational distribution over the latent sequence:
\begin{equation}\label{eq:localstep}
    q^*(\mb{x}) \propto \exp \left ( E_{q(A)} \left[  \ln \pi(x_1) \right] +
    \sum_{t=2}^T E_{q(A)} \left[  \ln A_{x_{t-1}, x_t} \right]  +\sum_{t=1}^T
    E_{q(\phi)} \left[  \ln p(y_t| x_t) \right] \right ).
 \end{equation} 
Compare with Eq.~\eqref{eq:HMMprob}.  Here, we have replaced probabilities by exponentiated expected log probabilities under the current variational distribution.  To determine the optimal $q^*(\mb{x})$ in Eq.~\eqref{eq:localstep}, define:
\begin{equation}\label{eq:modparam}
    \widetilde{A}_{j,k} := \exp\left[ E_{q(A)} \ln(A_{j,k}) \right] \quad  \widetilde{p}(y_t|x_t=k) := \exp\left[ E_{q(\phi)}
    \ln p(y_t|x_t=k) \right].
\end{equation}
We estimate $\pi$ with $\hat{\pi}$ being the leading eigenvector of $E_{q(A)}[A]$.  
We then use $\hat{\pi},\tilde{A}=(\widetilde{A}_{j,k})$, and
$\tilde{p}=\{\widetilde{p}(y_t|x_t=k), k=1,\ldots,K, t=1,\dots,T\}$ to run a forward-backward algorithm, producing forward messages $\alpha$ and backward messages
$\beta$ which allow us to compute $q^*(x_t=k)$ and $q^*(x_{t-1}=j,x_t=k)$.
\citep{Bishop:2006,Beale:2003}.  See the Supplement.

\subsection{Stochastic variational inference for non-sequential models}
\label{sec:svi}

Even in non-sequential models, the batch VB algorithm requires an entire pass through the dataset for each update of the global parameters.  This can be costly in large datasets, and wasteful when local-variable passes are based on uninformed initializations of the global parameters or when many data points contain redundant information.  

To cope with this computational challenge, \emph{stochastic variational
inference} (SVI) \cite{Hoffman:2013} leverages a Robbins-Monro algorithm
\cite{Robbins:Monro:1951} to optimize the ELBO via stochastic gradient ascent. 
When the data are independent, the ELBO in Eq.~\eqref{eq:elbo} can be expressed as 
\begin{equation}\label{eq:exchange}
\Ell = E_{q(\theta)} \left[ \ln p( \theta) \right] - E_{q(\theta)} \left[ \ln q( \theta ) \right]  + \sum_{i=1}^T E_{q(x_i)} \left[ \ln p( y_i, x_i |  \theta ) \right] - E_{q(\mb{x})} \left[ \ln q( \mb{x} )  \right]. 
\end{equation}
If a single observation index $s$ is sampled uniformly $s \sim \text{Unif}(1,
\ldots, T)$, the ELBO corresponding to $(x_s,y_s)$ as if it were replicated $T$ times is given by
\begin{equation}\label{eq:subexchange}
\Ell^s = E_{q(\theta)} \left[ \ln p( \theta) \right] - E_{q(\theta)} \left[ \ln q( \theta ) \right]  + T \cdot \left( E_{q(x_s)} \left[ \ln p( y_s, x_s |  \theta ) \right] - E_{q(x_s)} \left[ \ln q( x_s )  \right] \right),
\end{equation}
and it is clear that $E_{s}[\Ell^s] = \Ell$. At each iteration $n$ of the SVI algorithm, a
data point $y_s$ is sampled and its \emph{local} $q^*(x_s)$ is 
computed given the current estimate of global variational parameters $\mb{w}_n$. Next, the
\textit{global update} is performed via a noisy, unbiased gradient step ($E_s[\hat{\nabla}_{\mb{w}} \Ell^s] = \nabla_{\mb{w}} \Ell$). 
 When all pairs of
distributions in the model are conditionally conjugate, it is cheaper to
compute the stochastic \textit{natural gradient}, $\widetilde{\nabla}_{\mb{w}}
\Ell^s$, which additionally accounts for the information geometry of the distribution \cite{Hoffman:2013}. 
The resulting stochastic natural gradient step with step-size $\rho_n$ is:
\begin{equation}\label{eq:Rob-Monroe}
\mb{w}_{n+1} = \mb{w}_{n} + \rho_n \widetilde{\nabla}_{\mb{w}} \Ell^s (\mb{w}_{n}).
\end{equation}
We show the form of $\widetilde{\nabla}_{\mb{w}} \Ell^s$ in Sec.~\ref{sec:globalupdate}, specifically in Eq.~\eqref{eq:noisygrad} with details in the Supplement.

\section{Stochastic variational inference for HMMs}



The batch VB algorithm of
Sec.~\ref{sec:structuredMF} becomes prohibitively expensive as the length of
the chain $T$ becomes large.  
In particular, the forward-backward algorithm in the local step takes $O(K^2T)$
time.
Instead, we turn to a subsampling approach, but
naively applying SVI
from Sec.~\ref{sec:svi} fails in the HMM setting:  decomposing the sum over
local variables into a sum of independent terms as in
Eq.~\eqref{eq:exchange} ignores crucial transition counts, equivalent to
making a full mean-field approximation. 


Extending SVI to HMMs requires additional considerations due to the
dependencies between the observations.  It is clear that \emph{subchains} of consecutive observations rather than individual observations are necessary to capture the transition structure (see Sec.~\ref{sec:SVIELBO}).
We show that if the local variables of each subchain can be
exactly optimized, then stochastic gradients computed on subchains can be scaled to preserve unbiased estimates of the full gradient (see Sec.~\ref{sec:globalupdate}).


Unfortunately, as we show in Sec.~\ref{sec:local}, the local step becomes approximate due to edge effects: local variables are incognizant of nodes outside of the subchain during the forward-backward pass.
Although an \emph{exact} scheme requires message passing along the entire chain, we
harness the memory decay of the latent Markov chain to guarantee that local state
beliefs in each subchain form an \textit{$\epsilon$-approximation} $q_\epsilon(\mb{x})$ to the full-data beliefs $q^*(\mb{x})$.
We achieve these approximations by adaptively buffering the subchains with extra observations based on current global parameter estimates.  
We then prove that for $\epsilon$ sufficiently small, the noisy gradient computed using
$q_\epsilon(\mb{x})$ corresponds to an ascent direction in $\Ell$, guaranteeing convergence of our algorithm to a local optimum.  We refer to our algorithm, which is outlined in Alg.~\ref{alg:SVI}, as \emph{SVIHMM}.
\begin{algorithm}
   \caption{Stochastic Variational Inference for HMMs (SVIHMM)}
   \label{alg:SVI}
\begin{algorithmic}[1]
\STATE  Initialize variational parameters $(\mb{w}^A_0, \mb{w}^\phi_0)$ and choose stepsize schedule $\rho_n$, $n = 1, 2, \ldots$
\WHILE {(convergence criterion is not met)}
\STATE Sample a subchain $\mb{y}^S \subset \left\{ y_1, \ldots, y_T \right\}$ with $S \sim p(S)$
\STATE \textbf{Local step:} Compute $\hat{\pi},\widetilde{A},\widetilde{p}_S$ and run  $q(\mb{x}^S) = \texttt{ForwardBackward}(\mb{y}^S, \hat{\pi},\widetilde{A},\widetilde{p}_S)$.
\STATE \textbf{Global update:} 
$ \mb{w}_{n+1} = \mb{w_n}(1-\rho_n) + \rho_n (\mb{u} + \mb{c}^T E_{q(\mb{x}^S)}[t(\mb{x}^S,\mb{y}^S)] )$
\ENDWHILE
\end{algorithmic}
\end{algorithm}

\subsection{ELBO for subsets of data}
\label{sec:SVIELBO}
Unlike the independent data case (Eq.~\eqref{eq:exchange}), 
the local term in the HMM setting decomposes as
\begin{equation}\label{eq:local}
 \ln p(\mb{y,x} | \theta) = \ln \pi(x_1) + \sum_{t=2}^T \ln A_{x_{t-1}, x_t} +
 \sum_{i=1}^T \ln p(y_t| x_t).
 \end{equation} 
Because of the paired terms in the first sum, it is necessary to consider \textit{consecutive observations} to learn transition structure.
For the SVIHMM algorithm, we define our basic sampling unit as
\textit{subchains}  $\mb{y}^S= (y^S_1, \ldots, y^S_L)$, where $S$ refers to the
associated indices.
We denote the ELBO restricted to $\mb{y}^S$ as $\Ell^S$, and associated natural gradient as
$\widetilde{\nabla}_{\mb{w}}\Ell^S$.


\subsection{Global update}
\label{sec:globalupdate}

We detail the global update assuming we have optimized $q^*(\mb{x})$
\textit{exactly} (i.e., as in the batch setting), although this assumption will be relaxed as discussed in 
Sec~\ref{sec:local}. Paralleling Sec.~\ref{sec:svi}, the global SVIHMM step involves updating the global variational parameters $\mb{w}$ via stochastic (natural) gradient ascent based on $q^*(\mb{x}^S)$, the beliefs corresponding to our current subchain $S$. 

Recall from Eq.~\eqref{eq:subexchange} that the
original SVI algorithm maintains $E_s[\widetilde{\nabla}_{\mb{w}} \Ell^s]= \widetilde{\nabla}_{\mb{w}} \Ell$ by scaling the
gradient based on an individual observation $s$ by the total number of observations $T$.
In the HMM case, we analogously derive a \emph{batch factor} vector $\mb{c} = (c^A, c^\phi)$ such that 
\begin{equation}\label{eq:noisygrad} 
	E_S[\widetilde{\nabla}_{\mb{w}} \Ell^S]= \widetilde{\nabla}_{\mb{w}} \Ell \quad \mbox{with} \quad \widetilde{ \nabla}_{\mb{w}}  \Ell^S  = \mb{u} + \mb{c}^T E_{q^*(\mb{x}^S)} \left[ t(\mb{x}^S, \mb{y}^S) \right] -  \mb{w}. 
\end{equation}
The specific form of Eq. \eqref{eq:noisygrad} for Gaussian emissions is in the Supplement.
Now, the Robbins-Monro average in Eq.~\eqref{eq:Rob-Monroe} can be written as
\begin{equation}
    \label{eq:rminter}
    \mb{w}_{n+1} = \mb{w}_n(1-\rho_n) + \rho_n(\mb{u} +  \mb{c}^T E_{q^*(\mb{x}^S)}[t(\mb{x}^S,\mb{y}^S)] ).
\end{equation}
When the noisy natural gradients $\widetilde{\nabla}_{\mb{w}} \Ell^S$ are
independent and unbiased estimates of the true natural gradient, the iterates in
Eq.~\eqref{eq:rminter} converge to a local maximum of $\Ell$ under mild regularity
conditions as long as step-sizes $\rho_n$ satisfy $\sum_{n}
\rho_n^2 < \infty$, and $\sum_n \rho_n = \infty$ \cite{Bottou:1998,Hoffman:2013}. In our case, the noisy gradients are necessarily correlated even for independently sampled subchains due to dependence between observations $(y_1, \ldots, y_T)$.  However, as detailed in~\cite{Polyak:1973}, unbiasedness suffices for convergence of Eq.~\eqref{eq:rminter} to a local mode.
%
%

\paragraph{Batch factor}
Recalling our assumption of being at stationarity, $E_{q(\pi)}\ln \pi(x_1) =
E_{q(\pi)} \ln \pi(x_i)$ for all $i$. For a given subchain sampling rule $p(S)$ over subchains of length $L$, we can write 
\begin{equation}\label{eq:iterated}
E_{S} \bigg[ E_q \ln p(\mb{y}^S, \mb{x}^S | \theta)  \bigg] 
\approx p(S) E_q \left[ \sum_{t=1}^{T-L+1} \ln \pi(x_t) +
        (L-1) \sum_{t=2}^T \ln A_{x_{t-1},x_t} + L \sum_{t=1}^T p(y_t | x_t) \right] ,
\end{equation} 
where the expectation is with respect to $(\pi, A, \phi)$; this is detailed in the Supplement.
The approximate equality in Eq.~\eqref{eq:iterated} arises
because while most transitions appear in $L-1$ subchains, those near
the endpoints of the full chain do not, e.g., $x_1$ and $x_T$
appear in only one subchain. This error becomes negligible as the length of the HMM increases.
When $p(S)$ is uniform over all length $L$ subchains, by linearity of expectation the batch factor  $\mb{c} = (c^A, c^\phi)$ is given by $c^A = (T-L+1)/(L-1)$, $c^\phi = (T-L+1)/L$. Other choices of $p(S)$ can be implemented by iterated expectations analogously as in \cite{Gopalan:Mimno:Gerrish:Freedman:Blei:2012}, generally with a batch factor $\mb{c}^S$ varying with each subset $\mb{y}^S$.

\subsection{Local update}
\label{sec:local}

The optimal SVIHMM local variational distribution arises just as in the batch case of Eq. \eqref{eq:localstep}, but with time indices restricted to the length $L$ subchain $\mb{y}^S$:
\begin{equation}\label{eq:localSVI}
    q^*(\mb{x}^S) \propto \exp \left (  E_{q(A)} \left[  \ln \pi(x^S_1)\right] + \sum_{\ell=2}^L E_{q(A)} \left[  \ln A_{x^S_{\ell-1},x^S_\ell} \right] +\sum_{\ell=1}^L E_{q(\phi)} \left[  \ln p(y^S_\ell| x^S_\ell ) \right] \right ).
\end{equation}
To compute these local beliefs, we use our current $q(A), q(\phi)$---which have been informed by all previous subchains---to form $\hat{\pi}$, $\widetilde{A}$, $\widetilde{p}_S = \{\widetilde{p}(y^S_\ell | x^S_\ell = k), \forall k, \ell=1,\dots,L\}$, with these parameters defined as in the batch case. We then use these parameters in a forward-backward algorithm detailed in the Supplement.  However, this message passing produces only an approximate optimization due
to loss of information incurred at the ends of the subchain.
Specifically, for $\mb{y}^S = (y_t,\ldots,y_{t+L})$, the forward messages coming from
$y_1,\ldots,y_{t-1}$ are not available to $y_t$, and similarly the backwards
messages from $y_{t+L+1},\ldots,y_T$ are not available to $y_{t+L}$. 

Recall our assumption in the global update step that $q^*(\mb{x}^S)$ corresponds to a subchain of the full-data optimal beliefs $q^*(\mb{x})$.  Here, we see that this assumption is assuredly false; instead, we analyze the implications of using approximate local subchain beliefs and aim to ameliorate the edge effects.

\paragraph{Buffering subchains}
To cope with the subchain edge effects, we augment the subchain
$S$ with enough extra observations on each end so that the local state beliefs,
$q(x_i)$, $i \in S$, are within an $\epsilon$-ball of $q^*(x_i)$ --- those had we
considered the entire chain.  The practicality of this approach arises from the approximate finite memory of the process. In particular, consider performing a
forward-backward pass on $(x_{1-\tau}^S,\dots,x_{L+\tau}^S)$ leading to
approximate beliefs $\tilde{q}^{\tau}(x_i)$. Given $\epsilon > 0$, define $\tau_\epsilon$ as the smallest buffer length $\tau$ such that
\begin{align}
    \label{eq:epsapprox}
\max_{i \in S} || \tilde{q}^{\tau}(x_i) - q^*(x_i) ||_1 \leq \epsilon.
\end{align}
The $\tau$ that satisfies Eq.~\eqref{eq:epsapprox} determines the number
of observations used to \textit{buffer} the subchain.  After improving subchain beliefs, we discard $\tilde{q}^{\tau}(x_i)$, $i \in \mbox{\textit{buffer}}$, prior to the global update.  As will be seen in Sec.~\ref{sec:exper}, in practice the necessary $\tau_\epsilon$ is typically very small relative to the lengthy observation sequences of interest.

Buffering subchains is related to \textit{splash belief propagation}
(BP) for parallel inference in undirected graphical models, where the
belief at any given node is monitored based on locally-aware message passing
in order to maintain a good approximation to the true belief
\cite{Gonzalez:Low:Guestrin:2009}.
Unlike splash BP, we
embed the buffering scheme inside an iterative procedure for updating both the
local latent structure and the global parameters, which affects the
$\epsilon$-approximation in future iterations.
Likewise, we wish to maintain the approximation on an entire subchain, not just at a single node. 

Even in settings where parameters $\theta$ are known, as in splash BP, analytically choosing $\tau_\epsilon$ is generally infeasible. As such, we follow the approach of splash BP to select an approximate $\tau_\epsilon$. We then go further by showing that 
SVIHMM still converges using approximate messages
within an uncertain parameter setting where $\theta$ is learned simultaneously with the state sequence $\mb{x}$.

Specifically, we approximate $\tau_\epsilon$ by monitoring the change in belief residuals with a sub-routine \texttt{GrowBuf}, outlined in Alg.~\ref{alg:gb}, that iteratively expands a buffer $q^\mathrm{old} \rightarrow q^\mathrm{new}$ around a given subchain $\mb{y}^S$. \texttt{Growbuf} terminates when all belief residuals satisfy
\begin{align}
    \max_{i \in S} || q(x_i)^{\mathrm{new}} - q(x_i)^{\mathrm{old}} ||_1 \leq \epsilon.
\end{align}



The \texttt{GrowBuf} sub-routine can be computed efficiently due to (1)
monotonicity of the forward and backward messages 
so that only residuals at endpoints, $q(x^S_1)$ and
$q(x^S_L)$, need be considered, and (2) the reuse of computations.  Specifically, the forward-backward pass can be
rooted at the midpoint of $\mb{y}^S$ 
so that messages to the endpoints can be efficiently propagated, and vice versa~\cite{Russell:2003}.  

Furthermore, choosing sufficiently small $\epsilon$ guarantees that the noisy
natural gradient lies in the same half-plane as the true natural gradient, a sufficient condition for maintaining convergence when using approximate gradients \cite{Nocedal:Wright:2006}; the proof is presented in the Supplement.

\begin{algorithm}[h!]
   \caption{\texttt{GrowBuf} procedure.}
   \label{alg:gb}
\begin{algorithmic}[1]
\STATE \textbf{Input:} subchain $S$, min buffer length $u \in \mathbb{Z}_+$, error tolerance $\epsilon > 0$.
\STATE Initialize $q^{\mathrm{old}}(\mb{x}^S) = \texttt{ForwardBackward}(\mb{y}^S, \hat{\pi},\widetilde{A},\widetilde{p}_S)$ and set $S^{\mathrm{old}} = S$.
\WHILE { true }
\STATE Grow buffer $S^{\mathrm{new}}$ by extending $S^{\mathrm{old}}$ by $u$ observations in each direction.
\STATE $q^{\mathrm{new}}(\mb{x}^{S^{\mathrm{new}}}) = \texttt{ForwardBackward}(\mb{y}^{S^{\mathrm{new}}},\hat{\pi},\widetilde{A},\widetilde{p}_{S^\mathrm{new}})$, reusing messages from $S^\mathrm{old}$.
\IF {$\norm{q^{\mathrm{new}}(\mb{x}^{S}) - q^{\mathrm{old}}(\mb{x}^{S})} < \epsilon$}
\RETURN $q^*(\mb{x}^S) = q^{\mathrm{new}}(\mb{x}^{S}) $
\ENDIF
\STATE Set $S^\mathrm{old} = S^\mathrm{new}$ and $q^\mathrm{old} = q^\mathrm{new}$.
\ENDWHILE
\end{algorithmic}
\end{algorithm}

\subsection{Minibatches for variance mitigation and their effect on computational complexity}

Stochastic gradient algorithms often benefit from sampling multiple observations in order to reduce the variance of the gradient estimates at each
iteration.  We use a similar idea in SVIHMM by
sampling a \textit{minibatch} $B = (\mb{y}^{S_1}, \ldots, \mb{y}^{S_M})$ consisting of $M$
subchains.  If the latent Markov chain tends to dwell in one component
for extended periods, sampling one subchain may only contain
information about a select number of states observed in that component. 
Increasing the length of this subchain may only lead to redundant
information from this component. In contrast, using a minibatch of
many smaller subchains may discover disparate components of the
chain at comparable computational cost, accelerating learning and leading to a
better local optimum. However, subchains must be sufficiently long to be informative of transition dynamics.  In this setting, the local step on each
subchain is identical; summing over subchains in the minibatch yields
the gradient update:
\begin{equation*}
    \hat{\mb{w}}^B = \sum_{S \in B} \mb{c}^T E_{q(\mb{x}^S)} \left[ t(\mb{x}^S, \mb{y}^S) \right], \;\;\; \mb{w}_{n+1} =
    \mb{w}_n(1-\rho_n) + \rho_n \left ( u + \frac{\hat{\mb{w}}^B}{|B|}
    \right ).
\end{equation*}
We see that the computational complexity of SVIHMM is $O(K^2( L+2\tau_\epsilon) M)$, leading to significant efficiency gains compared to $O(K^2T)$ in batch inference when $( L+2\tau_\epsilon)M << T$.




\section{Experiments}
\label{sec:exper}

We evaluate the performance of SVIHMM compared to batch VB on synthetic experiments designed to illustrate the trade off between the choice of
subchain length $L$ and the number of subchains per minibatch $M$. We also
demonstrate the utility of \texttt{GrowBuf}.  We then apply our algorithm
to gene segmentation in a large human chromatin data set.
\paragraph{Synthetic data} We create two synthetic datasets with
$T=10,000$ observations and $K=8$ latent states.
The first, called \textit{diagonally dominant} (DD), illustrates the potential benefit of large $M$, the number of sampled subchains per minibatch. The Markov chain heavily self-transitions so that most subchains contain redundant information with observations generated from the same latent state.
Although transitions are rarely observed, the emission means are set to be
distinct so that
this example is likelihood-dominated and highly identifiable. Thus, fixing a
computational budget, we expect large $M$ to be preferable to large $L$,
covering more of the observation sequence and avoiding poor local modes arising
from redundant information.

The second dataset we consider contains two \textit{reversed cycles} (RC): the Markov chain strongly transitions from states $1 \to 2 \to 3 \to 1$ and $5
\rightarrow 7 \rightarrow 6 \rightarrow 5$ with a small probability of
transitioning between cycles via bridge states $4$ and $8$.
The emission means for the two cycles are very similar but occur in reverse
order with respect to the transitions. Transition information in observing long
enough dynamics is thus crucial to identify between states $1,2,3$ and $5,6,7$,
and a large enough $L$ is imperative.
The Supplement contains details for generating both synthetic datasets.

We compare SVIHMM to batch VB on these two synthetic examples.  For each per parameter setting, we ran 20 random restarts of SVIHMM for $100$ iterations and batch VB until convergence of the ELBO. A \textit{forgetting rate} $\kappa$ parametrizes step sizes $\rho_n = (1 + n)^{-\kappa}$. We fix the total number of observations $L \times M$ used per iteration of SVIHMM such that increasing $M$ implies decreasing $L$ (and vice versa).

\begin{figure}[!h]
\vspace{-10pt}
\centering
\subfigure[]{
    \includegraphics[scale=.705]{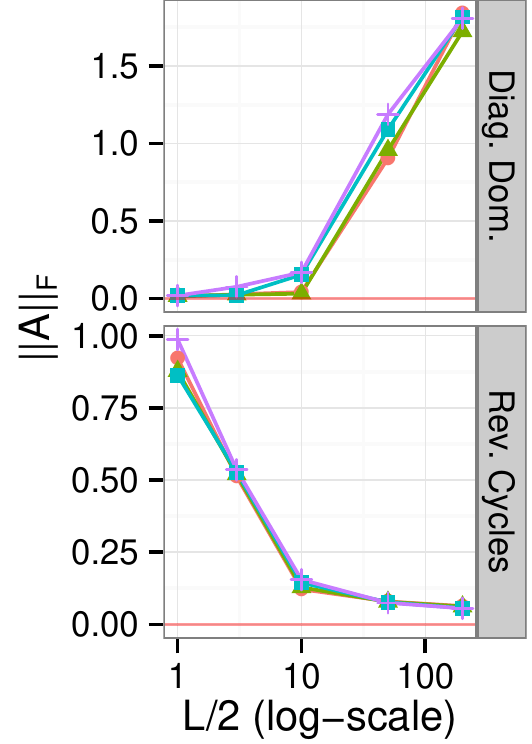}
    \label{fig:tran_synth}
}
\subfigure[]{
    \includegraphics[scale=.672]{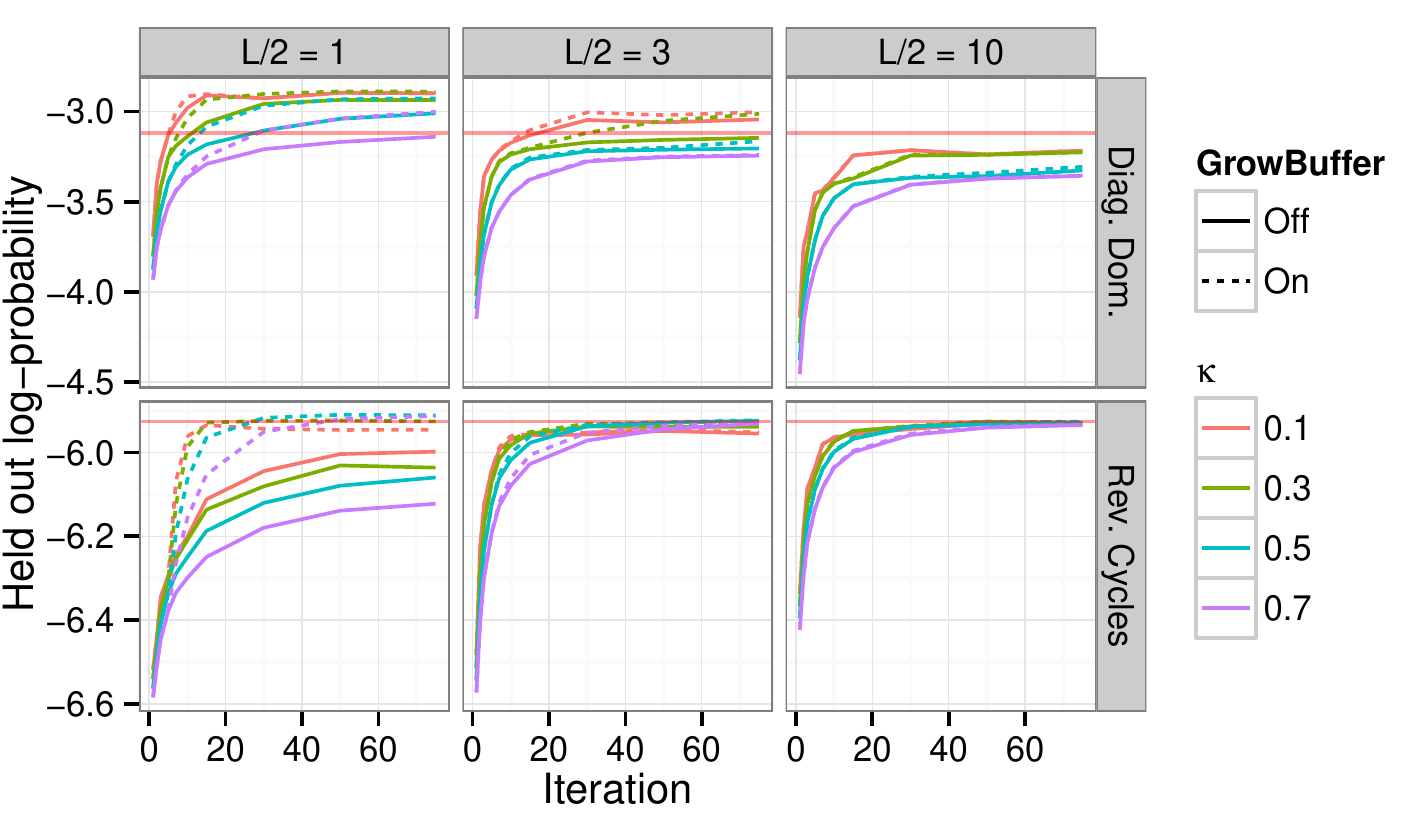}
    \label{fig:predlp}
}
\vspace{-10pt}
\caption{(a) Transition matrix error varying $L$ with $L \times M$ fixed. (b)
Effect of incorporating \texttt{GrowBuf}. Batch results denoted by horizontal red line in both figures.} \end{figure}

In Fig.~\ref{fig:tran_synth} we compare $||\hat{A} - A||_F$, where $A$ is the true transition matrix and $\hat{A}$ its learned variational mean. We see trends one would expect: the small $L$, large $M$ settings achieve better performance for the DD example, but the opposite holds for RC, with $\lfloor L/2 \rfloor=1$ significantly underperforming. (Of course, allowing large $L$ \emph{and} $M$ is always preferable, except computationally.) Under appropriate settings in both cases, we achieve comparable performance to batch VB. In Fig.~\ref{fig:predlp}, we see similar trends in terms of predictive log-probability holding out $10\%$ of the observations as a test set and using 5-fold cross validation.  Here, we actually notice that SVIHMM often achieves \emph{higher} predictive log-probability than batch VB, which is attributed to the fact that stochastic algorithms can find better local modes than their non-random counterparts.
%

A timing comparison of SVIHMM to batch VB with $T = 3$~million is presented
in Table~\ref{tab:timing}.  All settings of SVIHMM run faster than even a
single iteration of batch, with only a negligible change in predictive
log-likelihood. Further discussion on these timing results is in the Supplement.

\begin{table}
    \label{tab:timing}
    \caption{Runtime and predictive log-probability (without
    \texttt{GrowBuf}) on RC data.}
    \centering
    \begin{tabular}{|c|c|c|c|}
        \hline
        $\lfloor L/2 \rfloor$ & Runtime (sec.) & Avg. iter. time (sec.) & log-predictive \\
        \hline
        100 & $2.74 \pm 0.001$ & $0.03 \pm 0.000$ & $-5.915 \pm 0.004$ \\
        500 & $11.79 \pm 0.004$ & $0.12 \pm 0.000$ & $-5.850 \pm 0.000$ \\
        1000 & $23.17 \pm 0.006$ & $0.23 \pm 0.000$ & $-5.850 \pm 0.000$ \\
        batch & $1240.73 \pm 0.370$ & $248.15 \pm 0.074$ & $-5.840 \pm 0.000$ \\
        \hline
    \end{tabular}
\end{table}

Motivated by the demonstrated importance of choice of $L$, we now turn to examine the impact of the \texttt{GrowBuf} routine via predictive log-probability.
In Fig.~\ref{fig:predlp}, we see a noticeable improvement for small $L$ settings when \texttt{GrowBuf} is incorporated (the dashed lines in
Fig.~\ref{fig:predlp}). 
In particular, the RC example is now learning dynamics of the chain even with $\lfloor L/2 \rfloor= 1$, which was not possible without buffering. 
\texttt{GrowBuf} thus provides robustness by guarding against poor choice of $L$.
We note that the buffer routine does not overextend subchains, on average growing by only $\approx 8$
observations with $\epsilon = 1 \times 10^{-6}$.  Since the number of
observations added is usually small, \texttt{GrowBuf} does not significantly
add to per-iteration computational cost (see the Supplement).

\vspace{-.1in}
\paragraph{Human chromatin segmentation} We apply the SVIHMM algorithm to a massive human
chromatin dataset provided by the ENCODE project \cite{ENCODE:2012}.
This data was studied
in \cite{Hoffman:Buske:Wang:Weng:Bilmes:Noble:2012} with the goal of
unsupervised pattern discovery via \emph{segmentation} of the genome. Regions sharing the same labels have certain common properties in the observed data, and because the labeling at
each position is unknown but influenced by the label at the previous position,
an HMM is a natural model \cite{Day:Hemmaplardh:Thurman:Stamatoyannopoulos:Noble:2007}. 

We were provided with 250 million observations consisting of twelve assays carried out in the chronic myeloid leukemia cell line K562. We analyzed the data using SVIHMM on an HMM with 25 states and 12 dimensional Gaussian emissions. We compare our performance to the corresponding segmentation learned by an expectation maximization (EM) algorithm applied to a more flexible dynamic Bayesian network model (DBN) \cite{Hoffman:Ernst:Wilder:Kundaje:Harris:Libbrecht:2013}. Due to the size of the dataset, the analysis of~\cite{Hoffman:Ernst:Wilder:Kundaje:Harris:Libbrecht:2013} requires breaking the chain into several blocks, severing long range dependencies.

We assess performance by comparing the false discovery rate (FDR) of predicting
active promoter elements in the sequence.
The lowest (best) FDR achieved with SVIHMM over 20 random restarts trials was $.999026$ using $\lfloor L/2 \rfloor=2000, M = 50,\kappa = .5$\footnote{Other parameter settings were explored.}, comparable and slightly lower than the $.999038$ FDR obtained using DBN-EM on the severed data \cite{Hoffman:Ernst:Wilder:Kundaje:Harris:Libbrecht:2013}. We emphasize that even when restricted to a simpler HMM model, learning on the full data via SVIHMM attains similar results to that of \cite{Hoffman:Ernst:Wilder:Kundaje:Harris:Libbrecht:2013} with significant gains in efficiency. In particular, our SVIHMM runs require only under an hour for a fixed 100 iterations, the maximum iteration limit specified in the DBN-EM approach. In contrast, even with a parallelized implementation over the broken chain, the DBN-EM algorithm 
can take days.  In conclusion, SVIHMM enables scaling to the entire dataset, allowing for a more principled approach by utilizing the data jointly.
%
%

\section{Discussion}

We have presented stochastic variational inference for HMMs, extending such algorithms from
independent data settings to handle time dependence.  We elucidated the complications
that arise when sub-sampling dependent observations and proposed a
scheme to mitigate the error introduced from breaking dependencies.  
Our approach provides an adaptive technique with provable guarantees for
convergence to a local mode.
Further extensions of the algorithm in the HMM setting include adaptively selecting the length of
meta-observations and parallelizing the local step when the number of
meta-observations is large. Importantly, these ideas generalize to other settings and can be applied to Bayesian
nonparametric time series models, general state space models, and other graph structures with spatial dependencies.

\subsection*{Acknowledgements}
{\small This work was supported in part by the TerraSwarm Research Center sponsored by MARCO and DARPA, DARPA Grant FA9550-12-1-0406 negotiated by AFOSR, and NSF CAREER Award IIS-1350133. JX was supported by an NDSEG fellowship.  We also appreciate the data, discussions, and guidance on the ENCODE project provided by Max Libbrecht and William Noble.}

\renewcommand{\thetable}{S\arabic{table}}%
\renewcommand{\thefigure}{S\arabic{figure}}%
\renewcommand{\thesection}{S\arabic{section}}%
\renewcommand{\theequation}{S\arabic{equation}}%
\setcounter{section}{0}
\setcounter{figure}{0}
\setcounter{table}{0}
\setcounter{equation}{0}
\setcounter{footnote}{0}

\title{Supplemental Material:  Stochastic Variational Inference for Hidden Markov Models}
\maketitle

\vspace{-0.6in}
\section{Introduction}

In this document we present further details into the how to compute the
quantities necessary for the SVIHMM algorithm.  We also derive key equations
necessary for the analysis of the algorithm, and present and prove the
convergence theorem for stochastic gradient ascent using approximate noisy
natural gradients.  We then present specifics of the synthetic data that we
use to evaluate SVIHMM.  Last, we discuss the timing experiment in depth.

\section{Model specification and variational approximation}

Recall our model specification for a hidden Markov model with $K$ latent
states, Gaussian emissions $y_t \in \reals^p$, and conjugate Dirichlet and
normal-inverse-Wishart (NIW) priors on the rows of the transition matrix and
emission parameters, respectively.  Specifically, let $\alpha \in \reals_+^K$,
$\mu_0 \in \reals^p$, $\Sigma_0 \in \mathbb{S}^p_{++}$ a symmetric positive
definite matrix, $\kappa_0 > 0$, and $\nu_0 > p+2$.  Then, the model is
specified as:
\begin{equation}
\begin{aligned}
  A_{k:} &\sim \Dir(\mb{\alpha}), \;\; k = 1,\ldots,K \\
  \phi_k = (\mu_k,\Sigma_k) &\sim \NIW(\mu_0, \kappa_0, \Sigma_0, \nu_0), \;\; k = 1,\ldots,K \\
  x_1 | \pi_0 &\sim \Mult(\pi_0) \\
  x_t | x_{t-1} &\sim \Mult(A_{x_{t-1}}) \\
  y_t | x_t, \{\phi_k\}_{k=1}^K &\sim \N(\mu_{x_t}, \Sigma_{x_t}), \;\; t = 1,\ldots,T.
\label{eqn:model}
\end{aligned}
\end{equation}

The algorithms presented in the main paper use the natural parameterization of
the Dirichlet and NIW distributions which we provide here.  The natural
parameters of a $\Dir(\alpha)$ distribution are given by
$\mathbf{u}^A = \alpha -1 \in \reals^K$.  The natural parameters for the
$\NIW(\mu_0, \Sigma_0, \kappa_0, \nu_0)$ are denoted $\mathbf{u}^\phi =
(u^\phi_1, u^\phi_2, u^\phi_3, u^\phi_4)$ where the components are given by
\begin{equation}
    \begin{aligned}
        u^\phi_1 &= \kappa_0 \mu_0 \\
        u^\phi_2 &= \kappa_0 \\
        u^\phi_3 &= \Sigma_0 + \kappa_0 \mu_0 \mu_0^T \\
        u^\phi_4 &= \nu_0 + 2 + p.
\end{aligned}
    \label{eqn:niw_natpars}
\end{equation}

In the HMM model in Eq.~\eqref{eqn:model} each row of $A$ is given a
$\Dir(\alpha)$ prior so that there is a natural parameter for each row,
$\mathbf{u}^A_k \in \reals^K$.  Similarly, there is a natural parameter
corresponding to each emission distribution, $\mathbf{u}^\phi_k, k=1,\ldots,K$.

Recall from the main paper that we approximate the posterior of
Eq.~\eqref{eqn:model} as $p(A, \{\phi_k\}, \mathbf{x}) \approx
q(A)q(\{\phi_k\})q(\mathbf{x})$ governed by variational parameters
$\mathbf{w}^A$ and $\mathbf{w}^\phi$, respectively, where $q(A)$ is a product of
Dirichlet distributions (one per row of $A$) and $q(\{\phi_k\})$ is a product of
NIW distributions (one per emission distribution).
The variational distribution over the local variables, $q(\mathbf{x})$, is
represented by a $T \times K$ row stochastic matrix where the entry in row $t$
and column $k$ is $q(x_t=k)$.
We describe how to compute $q(\mathbf{x})$ in Sec.~\ref{sec:forward_backward}
of the Supplement.

\section{Expected sufficient statistics for a HMM with Gaussian emissions}
\label{sec:expected_suffstats}

As shown in the main paper, in order to perform batch VB~(Eq.~(6)) via
coordinate-ascent or
SVI~(Eq.~(14)) via stochastic gradient ascent on the model in
Eq.~\eqref{eqn:model}, we must be able to compute
the \textit{sufficient statistics}, $t(\cdot)$, of the various distributions.  In this
section we derive the necessary sufficient statistics for the HMM with Gaussian
emissions and conjugate priors described above~\cite{Beale:2003}.

In the batch setting, the sufficient statistics for the $j$th row of $A$ are
given by the number of transitions from state $j$ to each other state over the
entire observation sequence.  In particular, the sufficient statistics
corresponding to the transition from state $j$ to $k$ are given by:
\begin{equation}
    t^A_{jk}(\mathbf{x}) = \sum_{t=2}^T \ind_{x_{t-1}=j, x_t=k},
    \label{eqn:dir_suffstats_batch}
\end{equation}
where the indicator function $\ind_A$ is $1$ when event $A$ occurs, and $0$
otherwise.  Note that the sufficient statistics for the rows of the transition
matrix only depend on the latent state sequence and not on the actual
observations.  We then combine all sufficient statistics for the $j$th row into
the vector of counts
$t^A_j(\mathbf{x}) = (t^A_{j1}(\mathbf{x}),\ldots,t^A_{jK}(\mathbf{x}))$.  In the
main paper we suppress the $j$ notation, however, the update for each row of
$A$ uses the sufficient statistics corresponding to that row.

For the SVI case where we only consider a subchain of observations, $S$,
the sufficient statistics for the transition from state $j$ to $k$ is
given by:
\begin{equation}
    t^A_{jk}(\mathbf{x}) = \sum_{\ell=2}^L \ind_{x^S_{\ell-1}=j, x^S_\ell=k}.
    \label{eqn:dir_suffstats_sub}
\end{equation}
That is, we consider the number of times a transition from state $j$ to $k$
occurs in $S$ ignoring the rest of the observations.

To compute both the batch VB and SVI updates for the emission distributions we
need to compute the sufficient statistics of the NIW distribution.  Recall that
the natural parameterization of the NIW distribution corresponding to emission
$k$ is of the form $\mathbf{u}^\phi_k = (u^\phi_{k,1}, u^\phi_{k,2},
u^\phi_{k,3}, u^\phi_{k,4})$.  There will be a sufficient statistic
corresponding to each entry of $\mathbf{u}^\phi_k$, which in the batch setting
are given by:
\begin{equation}
    \begin{aligned}
        t^\phi_{k,1}(\mathbf{x}, \mathbf{y}) &= \sum_{t=1}^T y_t \ind_{x_t=k} \\
        t^\phi_{k,2}(\mathbf{x}, \mathbf{y}) &= \sum_{t=1}^T \ind_{x_t=k} \\
        t^\phi_{k,3}(\mathbf{x}, \mathbf{y}) &= \sum_{t=1}^T y_t y_t' \ind_{x_t=k} \\
        t^\phi_{k,4}(\mathbf{x}, \mathbf{y}) &= \sum_{t=1}^T \ind_{x_t=k}.
    \end{aligned}
    \label{eqn:niw_suffstats_batch}
\end{equation}
These sufficient statistics are identical to those obtained for a NIW prior
for independent Gaussian observations since conditioned on the state sequence,
$\mathbf{x}$, the observations are independent.  As above, the analogous NIW
sufficient statistics for a subchain, $S$, are given by:
\begin{equation}
    \begin{aligned}
        t^\phi_{k,1}(\mathbf{x}, \mathbf{y}) &= \sum_{\ell=1}^L y^S_\ell
        \ind_{x^S_\ell=k} \\
        t^\phi_{k,2}(\mathbf{x}, \mathbf{y}) &= \sum_{\ell=1}^L \ind_{x^S_\ell=k} \\
        t^\phi_{k,3}(\mathbf{x}, \mathbf{y}) &= \sum_{\ell=1}^L
        y^S_\ell (y^{S}_{\ell})' \ind_{x_\ell=k} \\
        t^\phi_{k,4}(\mathbf{x}, \mathbf{y}) &= \sum_{\ell=1}^L \ind_{x^S_\ell=k}.
    \end{aligned}
    \label{eqn:niw_suffstats_sub}
\end{equation}

For both the batch VB and SVI algorithms we need to compute the expectations of
the sufficient statistics with respect to the variational distribution
$q(\mathbf{x})$ which by Eqs.~\eqref{eqn:dir_suffstats_batch} and
\eqref{eqn:niw_suffstats_batch} are given by:
\begin{equation}
    \begin{aligned}
        \E_{q(\mathbf{x})}[t^A_{jk}(\mathbf{x})] &= \sum_{t=2}^Tq(x_{t-1} = j, x_t = k) \\
        \E_{q(\mathbf{x})}[t^\phi_{k,1}(\mathbf{x},\mathbf{y})] &= \sum_{t=1}^T y_t q(x_t =k) \\
        \E_{q(\mathbf{x})}[t^\phi_{k,2}(\mathbf{x},\mathbf{y})] &= \sum_{t=1}^T q(x_t =k) \\
        \E_{q(\mathbf{x})}[t^\phi_{k,3}(\mathbf{x},\mathbf{y})] &= \sum_{t=1}^T y_ty_t' q(x_t=k) \\
        \E_{q(\mathbf{x})}[t^\phi_{k,4}(\mathbf{x},\mathbf{y})] &= \sum_{t=1}^T q(x_t = k).
    \end{aligned}
    \label{eqn:E_suff_stats_batch}
\end{equation}
The expected sufficient statistics for a subchain $S$ are computed analogously,
restricting the computations in Eq.~\eqref{eqn:E_suff_stats_batch} to the
observations in the subchain.  In particular, they are computed as:
\begin{equation}
    \begin{aligned}
        \E_{q(\mathbf{x})}[t^A_{jk}(\mathbf{x})] &=
        \sum_{\ell=2}^Lq(x^S_{\ell-1} = j, x^S_\ell = k) \\
        \E_{q(\mathbf{x})}[t^\phi_{k,1}(\mathbf{x},\mathbf{y})] &=
        \sum_{\ell=1}^L y_t q(x^S_\ell =k) \\
        \E_{q(\mathbf{x})}[t^\phi_{k,2}(\mathbf{x},\mathbf{y})] &=
        \sum_{\ell=1}^L q(x^S_\ell =k) \\
        \E_{q(\mathbf{x})}[t^\phi_{k,3}(\mathbf{x},\mathbf{y})] &=
        \sum_{\ell=1}^L y_ty_t' q(x^S_\ell=k) \\
        \E_{q(\mathbf{x})}[t^\phi_{k,4}(\mathbf{x},\mathbf{y})] &=
        \sum_{\ell=1}^L q(x^S_\ell = k).
    \end{aligned}
    \label{eqn:E_suff_stats_sub}
\end{equation}

We can then plug the expected sufficient statistics into Eqs.~(6) or (14) in
the main paper to determine coordinate-ascent or stochastic gradient updates,
respectively.
However, in order to compute the expected sufficient statistics in either the
coordinate-ascent (batch VB) or stochastic gradient-ascent (SVI) algorithms we
must first
compute $q(\mathbf{x})$ for batch VB or $q(\mathbf{x}^S)$ for SVI.  We describe
how to do this in the next section.

\section{Forward-backward algorithm for local variational update}
\label{sec:forward_backward}

The optimal distribution over the local variables, $q^*(\mathbf{x}) =
q^*(x_1,\ldots,x_T)$ for batch VB and $q^*(x^S) = q^*_(x^S_1,\ldots,x^S_L)$
for SVI,
is needed in order to compute the expected sufficient statistics that appear in
the coordinate-ascent and gradient equations for the global parameters.  In
particular, looking at Eq.~\eqref{eqn:E_suff_stats_batch} we need to be able to 
compute the \textit{marginal-beliefs} of each hidden state, i.e.\ $q^*(x_t)$, and
the \textit{pairwise-beliefs}, $q^*(x_{t-1}, x_t)$.
Following~\cite{Beale:2003}
we use the forward-backward algorithm, a dynamic programming algorithm, to
determine the marginal- and pairwise-beliefs in time $O(K^2T)$.

Recall Eq.~(7) from the main paper which describes the form of the optimal
variational distribution for the local parameters:
\begin{equation}
    \label{eqn:local_opt}
    q^*(\mb{x}) \propto \exp \left ( E_{q(A)} \left[  \ln \pi(x_1) \right] +
    \sum_{t=2}^T E_{q(A)} \left[  \ln A_{x_{t-1}, x_t} \right]  +\sum_{t=1}^T
    E_{q(\phi)} \left[  \ln p(y_t| x_t) \right] \right ).
\end{equation}
First, we define auxiliary parameters
\begin{equation}\label{eq:modparam}
    \widetilde{A}_{j,k} := \exp\left[ E_{q(A)} \ln(A_{j,k}) \right] \quad  \widetilde{p}(y_t|x_t=k) := \exp\left[ E_{q(\phi)}
    \ln p(y_t|x_t=k) \right]
\end{equation}
which we then use in the forward-backward algorithm as follows.  Note
$\widetilde{A} = (\widetilde{A}_{j,k})$ and $\widetilde{p}(y_t|x_t=k)$ can
be loosely interpreted as the expected sufficient statistics of the global
parameters.  For the HMM defined in Eq.~\eqref{eqn:model} we have that
\begin{align}
    \widetilde{A}_{j,k} &= \exp \left [ \psi \left (w^A_{jk} \right ) - \psi \left ( \sum_{l=1}^K
      w^A_{jl} \right ) \right ], \quad j,k \in 1,\ldots,K
\end{align}
where $\psi(\cdot)$ is the digamma function and
$\log\widetilde{p}(y_t|x_t=k, \phi)$ is given by the expectation under the
NIW variational distribution of the log-probability density of a Gaussian
distribution, the details of which can be found
in~\cite{Bishop:2006}(Ch. 10.2.1).

In the batch VB case we use the auxiliary parameters to propagate a set of
\textit{forward messages},
$\alpha = (\alpha_{t,k}), t\in 1,\ldots,T, k\in 1,\ldots,K$, starting at $t=1$
according to:
\begin{equation}
    \alpha_{1,k} = \pi_{0,k}, \qquad \alpha_{t,k} = \sum_{j=1}^K \alpha_{t-1,j}
    \widetilde{A}_{j,k}\widetilde{p}(y_t|x_t=k),
\end{equation}
where $\pi_{0,k} = p(x_1 = k)$ is the initial distribution.  We then propagate a
set of \textit{backward messages},
$\beta = (\beta_{t,k}), t\in 1,\ldots,T, k\in 1,\ldots,K$, starting at $t=T$
and going backwards as:
\begin{equation}
    \beta_{T,k} = 1, \qquad \beta_{t,k} = \sum_{j=1}^K \widetilde{A}_{k,j}
    \widetilde{p}(y_{t+1}|x_{t+1}) \beta_{t+1,k}.
\end{equation}
The forward messages perform a filtering pass by propagating information
forwards in time, while the backwards messages perform a smoothing pass by
taking into account the information that future observations provide.  The use
of the auxiliary parameters is necessary since in Eq.~\eqref{eqn:local_opt} the
expectation and logarithm are not interchangeable.  For an in depth derivation
of the forward and backward recursions see~\cite{Beale:2003}.

Given the forward and backward messages we can compute the quantities of
$q^*(\mathbf{x})$ necessary for the global step.  In particular, the marginal
beliefs are given by
\begin{equation}
q^*(x_t =k) \propto \alpha_{t,k}\beta_{t,k}
\end{equation}
and the pairwise beliefs by
\begin{equation}
    q^*(x_{t-1}=j, x_t=k) \propto \alpha_{t-1,j}A_{j,k}p(y_t|x_t=k)\beta_{t,k}.
\end{equation}

For SVI, the forward-backward algorithm remains largely the same.  The major
difference is that only observations and local variables in the subchain are
considered.  The corresponding modifications to the above equations are
straight forward.  Additionally, since in the SVI setting we cannot learn the
initial state distribution, $\pi_0$, we initialize the forward messages as
$\alpha_{1,k} = \hat{\pi}_k$, where as described in the main paper, $\hat{\pi}$
is the leading eigenvector of $E_{q(A)}[A]$.

\section{Batch variational Bayes global udpate}

The batch VB global update for the model in Eq.~\eqref{eqn:model} is given by:
\begin{equation}
    \begin{aligned}
        \mathbf{w}^A_{jk} &= \mathbf{u}^A_{k} + \sum_{t=2}^T q(x_{t-1}=j,
        x_t=k), \quad j,k \in 1,\ldots,K \\
        w^\phi_{k,r} &= \mathbf{u}^\phi_{k,r} +
        \E_{q(\mathbf{x})}[t^\phi_{k,r}(\mathbf{x},\mathbf{y})], \quad k\in
        1,\ldots,K, \; r \in 1,\ldots,4
    \end{aligned}
\end{equation}
where the expectations with respect to $q(\mathbf{x})$ are given in
Eq.~\eqref{eqn:E_suff_stats_batch} and where quantities of $q(\mathbf{x})$ are
computed via the forward-backward algorithm described previously.  The index
$r$ indexes the sufficient statistics of the emission distributions, of which
there are four in the case of the NIW.

\section{Stochastic natural gradients for SVIHMM}

The natural gradients (Eq.~(14) in the main paper) for the model in
Eq.~\eqref{eqn:model} are given by:
\begin{equation}
\begin{aligned}
    \left[ \widetilde{\nabla}_{w^A} \Ell^S \right]_{jk} &=
        u_{jk}^A + c^A \sum_{\tau=2}^L q(x^S_{\tau-1}=j , x^S_\tau= k) -
        w_{jk}^A \\
        \left [ \widetilde{\nabla}_{w^{\phi_k}} \Ell^S \right ]_r &= u^\phi_r +
        c^\phi_r
        \sum_{\tau=1}^L E_{q(x^S)}[ t^{\phi}_{k,r}(x^S, y^S)] -
        w^\phi_{k,r}, \quad k \in 1,\ldots,K, \; r \in 1,\ldots,4,  .
    \label{eqn:natgrad}
\end{aligned}
\end{equation}
Quantities involving $q(\mb{x}^S)$ are computed using the
forward-backward algorithm in Sec.~\ref{sec:forward_backward} and the expected
sufficient statistics are derived in Sec.~\ref{sec:expected_suffstats}.
The gradients in Eq.~\eqref{eqn:natgrad} are then used in a Robbins-Monro
averaging procedure to update the global variational parameters.

%

\section{Batch factor}

As described in Sec.~3.2 of the main paper, in order to obtain
an unbiased estimate of the natural gradient of the $\Ell$ (Eq.~(12) in the
main paper) we must scale the terms of $\Ell^S$ to match the
size of the original data set.  Here we derive Eq.~(15) from the main paper
which allows us to read off the necessary factors to scale the natural
gradient.  As in the paper, we assume that a subchain, $S$, of length $L$ is
sampled according to $p(S) = \frac{1}{T-L+1}$ which results in:
\begin{equation}
\label{eq:iterated}
\begin{split}
E_{S} \bigg[ E_q \ln p(\mb{y}^S, & \mb{x}^S | \theta)  \bigg] =
\frac{1}{T-L+1}E_q  \bigg[  \ln \pi(x_1) + \sum_{t=2}^L \ln A_{x_{t-1}, x_t} +
\sum_{t=1}^L \ln p(y_t | x_t) \\
  &+ \ln \pi(x_2) + \sum_{t=3}^{L+1} \ln A_{x_{t-1}, x_t} + \sum_{t=2}^{L+1} \ln 
p(y_t | x_t) + \ldots \\
  &+ \ln \pi(x_{T-L+1}) + \sum_{t=T-L+2}^T \ln A_{x_{t-1}, x_t} + \sum_{t=T-L+1}^{T} 
  \ln p(y_t | x_t) \bigg] \\
  &  \approx \frac{1}{T-L+1} E_q \bigg [ \sum_{t=1}^{T-L+1} \ln \pi(x_t) +
        (L-1) \sum_{t=2}^T \ln A_{x_{t-1},x_t} + L \sum_{t=1}^T p(y_t | x_t) \bigg ] .
\end{split}
\end{equation} 
The approximation arises because the observations near the endpoints of the
observation sequence appear in fewer subchains than those in the middle of the
sequence, e.g.\ $x_1$ and $x_T$ only appear in one subchain.  However, the
error introduced from this approximation becomes
negligible as the length of the sequence increases which is the case we are
interested in.  From Eq.~\eqref{eq:iterated} we can read off the batch factors
as $\mathbf{c} = (c^A, c^\phi)$, where $c^A = (T-L+1)/(L-1)$, and
$c^\phi = (T-L+1)/L$.  More general choices for $p(S)$ may be used resulting in
different batch factors.

\section{Preservation of ascent direction with approximate local messages}


\begin{theorem}
If the noisy gradient with respect to the ``true" messages
\[ \hat{ \nabla}_{\mb{w}}  \Ell^S  = \mb{u} + E_{q^*} \left[ \mb{c}^T t(\mb{x}^S, \mb{y}^S) \right] -  \mb{w}
\]
lies in the same half plane as the noisy gradient with respect to approximate messages
\[
\bar{ \nabla}_{\mb{w}}  \Ell^S  = \mb{u} + E_{q_\epsilon} \left[ \mb{c}^T t(\mb{x}^S, \mb{y}^S) \right] -  \mb{w},
\]
then $\bar{\nabla}_{\mb{w}} \Ell^S$ is an ascent direction for $\Ell$ so that
SVIHMM will converge to a local maximum of the
ELBO~\cite{Polyak:1973,Nocedal:Wright:2006}.  To
ensure the gradients are in the same half-plane, it suffices to choose 
\[ \epsilon \leq \frac{ M^S(\mb{w}) }{ || \mb{c}^T t(\mb{x,y}) ||_2}, \] 
where  \[ M^S(\mb{w}) := \max \left (\norm{\hat{ \nabla}_{\mb{w}}\Ell^S
}_2,\norm{\bar{ \nabla}_{\mb{w}}  \Ell^S}_2 \right ) \]
\end{theorem}

\begin{proof}
    Let $\mb{y}^S = (y^S_1,\ldots,y^S_L)$ be a subchain of observations where
    $L << T$ and $\mb{x} = (x_1, \ldots, x_L)$ denote any configuration of
    latent states corresponding to $\mb{y}^S$.  Also assume we have an
    approximation $q_\epsilon(\mb{x})$ such that \[ \max_\mb{x} |q_\epsilon(\mb{x}) - q^*(\mb{x}) | < \epsilon \]
where $q^*(\mb{x})$ again denotes the ``true" distribution as if a full message
pass were performed on the entire dataset of length $T$. In our setting, $q^*$ is a discrete distribution (of dimension $K \times L$) over the latent state sequence, and $t$ is some $d$-dimensional sufficient statistic function that we assume is bounded. The proof follows analogously in the continuous case as long as $q^*$ and $q_\epsilon$ are absolutely continuous with respect to the same measure-- one simply substitutes the summations over $\mb{x}$ below with integration.

To show that $\bar{ \nabla}_{\mb{w}} \Ell^S$ lies in the same half-plane as  $\hat{ \nabla}_{\mb{w}}  \Ell^S$, it is sufficient that 
\[
\norm{\hat{ \nabla}_{\mb{w}}  \Ell^S -  \bar{ \nabla}_{\mb{w}}  \Ell^S }_2 <
\max \left (\norm{\hat{ \nabla}_{\mb{w}}\Ell^S }_2,\norm{\bar{ \nabla}_{\mb{w}}
\Ell^S}_2 \right ) \equiv M^S(\mb{w}). \]
Since 
$\mb{w}$ and $\mb{u}$ are independent of $q^*(\mb{x})$, we may translate the gradient vectors by $\mb{u} - \mb{w}$ and equivalently seek to show that 
\[
\norm{ E_{q_\epsilon} [ \mb{c}^T t(\mb{x},\mb{y}) ] - E_{q^*}[ \mb{c}^T
t(\mb{x},\mb{y}) ] }_2 < 
M^S(\mb{w}).
\]

Considering the difference component-wise, we have
\begin{align*}
\norm{E_{q_\epsilon} [ \mb{c}^T t(\mb{x},\mb{y}) ] - E_{q^*}[ \mb{c}^T
t(\mb{x},\mb{y}) ]}_2^2 &= \sum_{j=1}^d \left( c_j \sum_\mb{x} t_j(\mb{x,y}) q_\epsilon(\mb{x}) -  c_j \sum_\mb{x} t_j(\mb{x,y}) q^*(\mb{x}) ) \right)^2 \\
&=  \sum_{j=1}^d \left( c_j \sum_\mb{x} t_j(\mb{x,y}) ( q_\epsilon(\mb{x}) - q^*(\mb{x}) ) \right)^2 \\
& \leq \sum_{j=1}^d \left( c_j \sum_\mb{x} | t_j(\mb{x,y})|  | q_\epsilon(\mb{x}) - q^*(\mb{x}) | \right)^2 \\
& \leq \epsilon^2 \sum_{j=1}^d  \left( c_j \sum_{\mb{x}} | t_j(\mb{x,y}) |
    \right)^2 = \epsilon^2  \norm{\mb{c}^T t(\mb{x,y})}_2^2.
\end{align*}

Finally, since we want this quantity to be bounded above by
$M^S(\mb{w})^2$, we choose
\[
    \epsilon \leq \frac{ M^S(\mb{w}) } { \norm{\mb{c}^T t(\mb{x,y})}_2 }
\]
\end{proof}

As one would expect, ascent direction is preserved in the limit as $\epsilon
\rightarrow 0$ as long as $t(\cdot,\cdot)$ is a bounded sufficient statistic.
Also, we note that while the upper bound is not easy to evaluate to guide our choice of $\epsilon$ since true messages are unavailable, we show empirically that setting small values $\epsilon = 1 \times 10^{-6}$ in \texttt{GrowBuf} leads to noticeable performance gains empirically in the experiments section.

\begin{figure}[h]
\begin{center}
\subfigure[]{
    \framebox{
        \includegraphics[width=.312\textwidth]{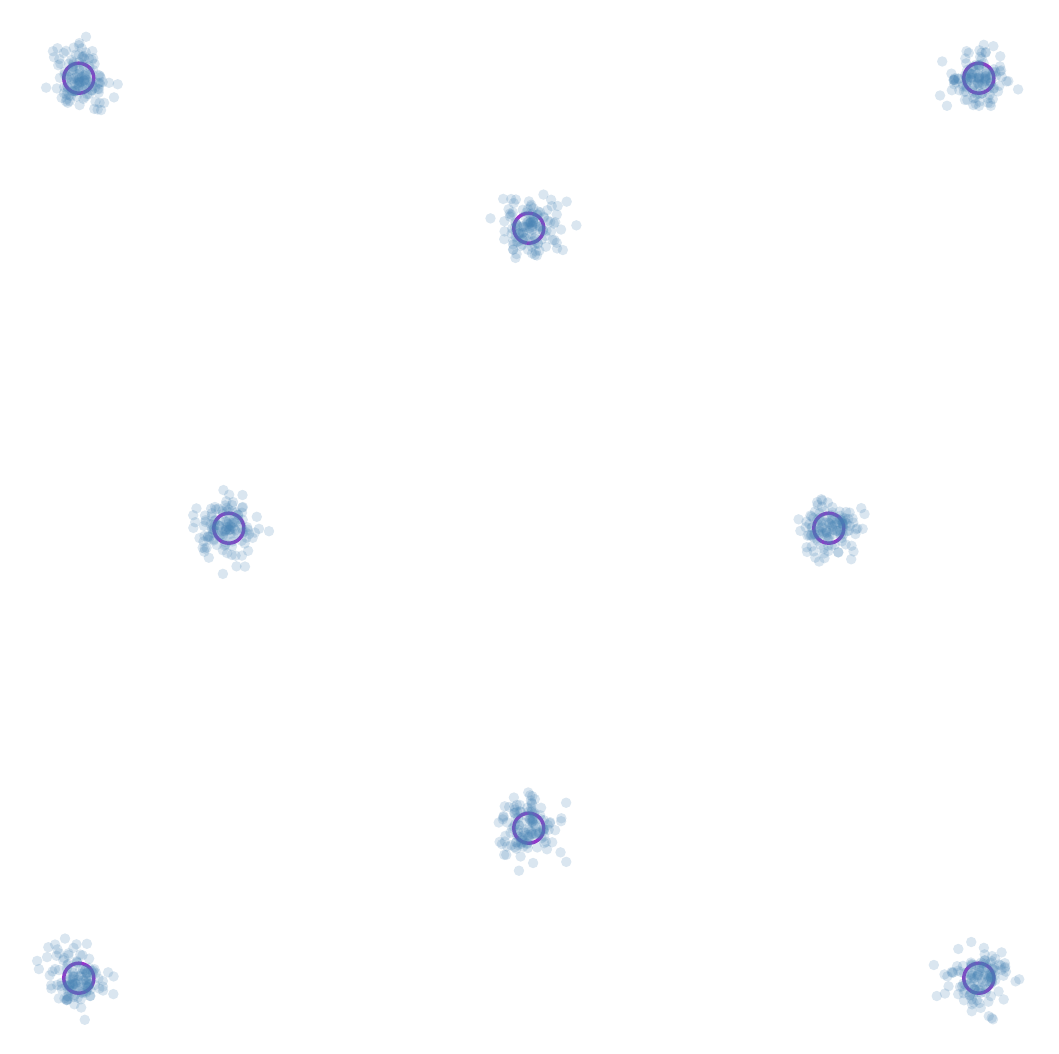}
    }
}
\subfigure[]{
    \framebox{
        \includegraphics[width=.306\textwidth]{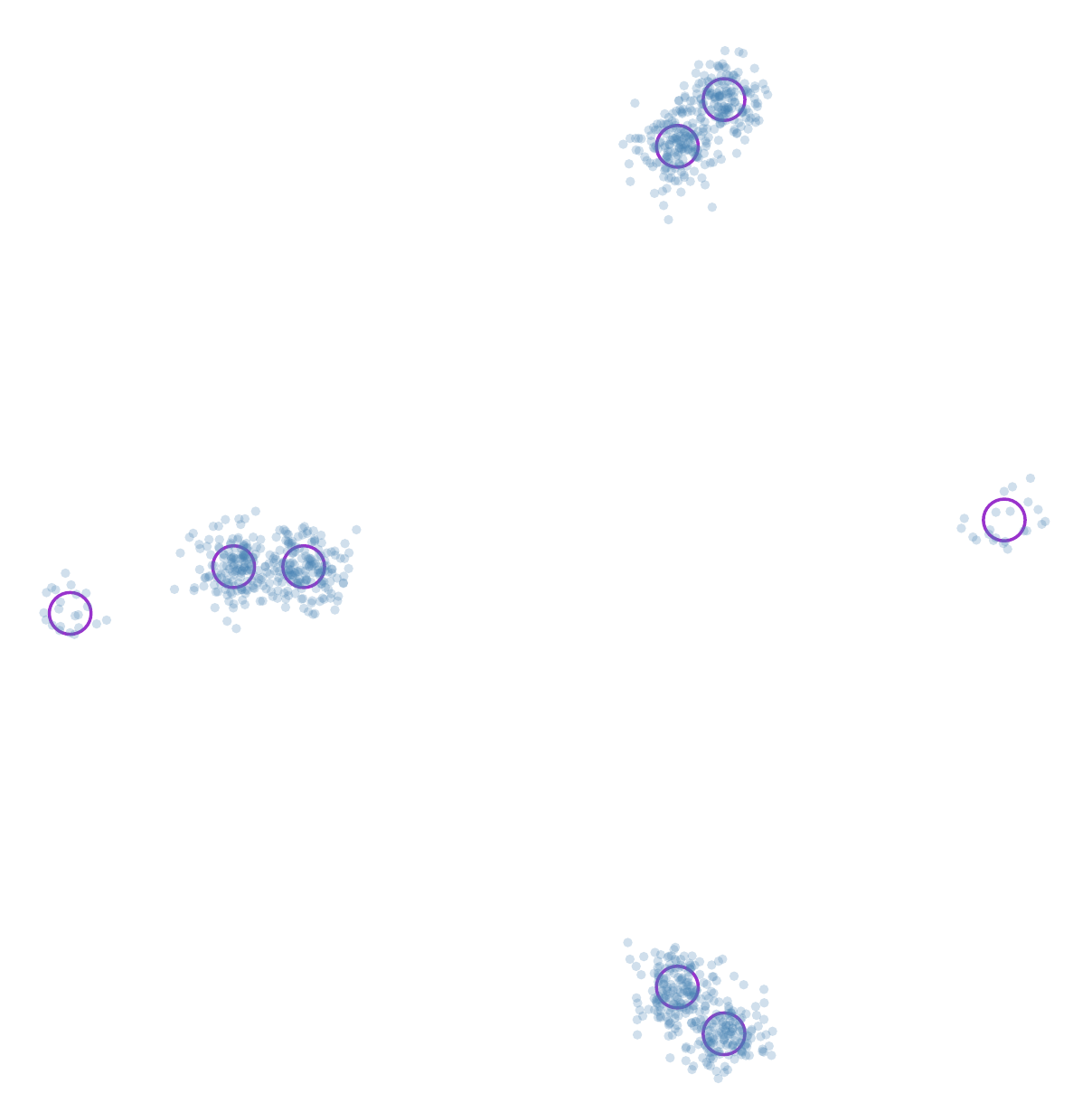}
    }
}
\caption{Observations generated from the diagonally dominant (left) and reversed 
cycles (right) examples.  Ellipses indicate true covariance matrices of underlying 
components.}
\label{fig:synth_data}
\end{center}
\end{figure}

\section{Synthetic data sets}

In this section we present the \textit{diagonally dominant} and
\textit{reversed cycles} synthetic data sets in detail.

The diagonally dominant data set uses the following transition matrix:
\[ A = \left( \begin{array}{cccccccc}
.999 & .001 & 0 & 0 & 0 & 0 & 0 & 0 \\
0 & .999 & .001 & 0 & 0 & 0 & 0 & 0 \\
0 & 0 & .999 & .001 & 0 & 0 & 0 & 0  \\
0 & 0 & 0 & .999 & .001 & 0 & 0 & 0 \\
0 & 0 & 0 & 0 & .999 & .001 & 0 & 0 \\
0 & 0 & 0 & 0 & 0 & .999 & .001 & 0 \\
0 & 0 & 0 & 0 & 0 & 0 & .999 & .001 \\
.001 & 0 & 0 & 0 & 0 & 0 & 0 & .999 
\end{array} \right) . \]
We see that there is a large probability that the observation sequence remains
in the same state.  The component means are given by
\begin{equation*}
\boldsymbol{\mu} = \left\{
(0,20); (20,0); (-90,-30); (30,-30); (-20,0); (0,-20); (30,30); (-30,30)
\right\},
\end{equation*}
where all component covariances are given by the $2 \times 2$ identity matrix,
$I_2$.
The emission distributions and simulated data are depicted in
Fig.~\ref{fig:synth_data} (left) and are meant to
be highly identifiable so that learning is largely
likelihood-dominated.  This illustrates the importance of sampling disparate
sections of the observation sequence in order for the global updates to contain
sufficient information to obtain accurate estimates.

The reversed cycles data set consists of
two 3-state cycles with essentially deterministic dynamics.  The two cycles are
connected by two bridge states that the process visits rarely to switch between
the cycles.  The state dynamics correspond to the following transition matrix:
\[ A = \left( \begin{array}{cccccccc}
.01 & .99 & 0 & 0 & 0 & 0 & 0 & 0 \\
9 & .01 & .99 & 0 & 0 & 0 & 0 & 0  \\
.85 & 0 & 0 & .15 & 0 & 0 & 0 & 0 \\
0 & 0 & 0 & 0 & 1 & 0 & 0 & 0 \\
0 & 0 & 0 & 0 & .01 & .99 & 0  & 0 \\
0 & 0 & 0 & 0 & 0 & .01 & .99 & 0 \\
0 & 0 & 0 & 0 & .85 & 0 & 0 & .15 \\
1 & 0 & 0 & 0 & 0 & 0 & 0 & 0
\end{array}\right) . \]
The emission means are set to
\begin{equation*}
\boldsymbol{\mu} = \left\{ (-50,0); (30,-30); (30,30); (-100,-10); (40,-40); (-65,0); (40, 40); (100,10)  \right\},
\end{equation*}
with covariance matrices given by $20*I_2$.
Observations generated from this model and the emission distributions are
shown in Fig.~\ref{fig:synth_data} (right).
The means of emissions $1$ and $5$, states $2$ and $6$, and states $3$ and $7$
have indistinguishable means, but the cycles $1\rightarrow 2 \rightarrow 3$ and
$5 \rightarrow 6 \rightarrow 7$ visit the means in reverse orders.  The
emission means of the bridge states are far from the two cycles so that they
are identifiable.  Learning the transition dynamics in this case is key in
order to learn the overlapping emissions.

\section{Discussion of timing experiment}

Here we explain our choice of settings for the timing comparison between SVIHMM
and batch VB in Sec.~4 of the main paper.  We implemented both the SVIHMM and
batch VB algorithms in Python except that the forward-backward algorithm was
written in C++.
Additionally, since SVIHMM operates on shorter sequences than batch VB it
does not benefit as much from the optimized forward-backward algorithm.
The gradient computations for SVIHMM were not optimized and are subject to
Python overhead, however, the coordinate-ascent update for bath VB are
vectorized using Numpy.
Therefore, in order to compare the batch VB and SVIHMM algorithms fairly we
set $T = 3$ million and $M =1$ as increasing $M$ results in higher overhead due
to the interpreted nature of Python
which could be mitigated in C++.  Since $M$ is small,
$L$ must be
chosen relatively large in order to obtain informative gradients.  For large
$L$ the \texttt{growBuf} routine negligibly affects the predictive
log-likelihood and the running time of the algorithm since the length of the
subchain causes the message
error to be small and thus few observations are added as a buffer.

\small{
\bibliography{SVINIPS}
\bibliographystyle{unsrt}
}

\end{document}